\documentclass[letterpaper, 10 pt, conference]{IEEEtran}
\IEEEoverridecommandlockouts

\usepackage{graphicx}
\graphicspath{ {./Figures/} }
\IEEEoverridecommandlockouts
\usepackage[american]{babel}
\usepackage[utf8]{inputenc}
\usepackage[T1]{fontenc}
\usepackage{newfloat}
\usepackage{booktabs}
\usepackage{amssymb, amsmath, array, bm, algorithm, standalone,csquotes,textcomp, amssymb,amsfonts, tabularx}
\usepackage{gensymb, wasysym} 
\usepackage{graphicx, xcolor, subcaption, soul}
\usepackage{siunitx, nameref, zref-xr}
\usepackage{algorithmic}
\usepackage{comment}
\usepackage[]{mdframed}
\zxrsetup{toltxlabel}
\usepackage{microtype}
\usepackage[nolist]{acronym}
\usepackage{hyperref}
\usepackage{multirow}
\usepackage[table]{xcolor}
\usepackage{amssymb}
\usepackage{amssymb}%
\usepackage{pifont}%

\definecolor{myBlue}{HTML}{03456A} 
\definecolor{Gray}{RGB}{230,230,230}

\usepackage{makecell, tabularx}
\newcolumntype{L}{>{\raggedright\arraybackslash}X}
\usepackage{siunitx}
\usepackage{booktabs}

\hypersetup{
    colorlinks=true,
    linkcolor=myBlue,
    citecolor=myBlue,
    filecolor=myBlue,      
    urlcolor=myBlue,
    pdftitle={main},
    pdfpagemode=FullScreen,
    }
\usepackage[nameinlink, capitalise]{cleveref}

\makeatletter %
\newcommand{\linebreakand}{%
  \end{@IEEEauthorhalign}
  \hfill\mbox{}\par
  \mbox{}\hfill\begin{@IEEEauthorhalign}
}
\makeatother%

\newcommand\etal[0]{\textit{et\,al.\ }}
\newcommand\mailto[1]{\href{mailto:#1}{#1}}

\newcommand{\SuccessCell}{\color{green}\checkmark}
\newcommand{\FailureCell}{\color{red}\ding{55}}
\newcommand{\InvalidCell}{\textcolor{gray}{Inv}}

\let\oldcite\cite
\renewcommand*\cite[1]{\,\oldcite{#1}}
\def\BibTeX{{\rm B\kern-.05em{\sc i\kern-.025em b}\kern-.08em
    T\kern-.1667em\lower.7ex\hbox{E}\kern-.125emX}}

\begin{document}
\begin{acronym}
    \acro{DMS}{distributed manipulator systems}
    \acro{DoF}{degrees of freedom}
    \acro{PLA}{polylactic acid}
    \acro{PET}{polyethylene terephthalate}
\end{acronym}

\title{Flexible and Foldable: Workspace Analysis and Object Manipulation Using a  Soft, Interconnected, Origami-Inspired Actuator Array \\
}

\author{
    \IEEEauthorblockN{
    Bailey Dacre\IEEEauthorrefmark{1},
    Rodrigo Moreno\IEEEauthorrefmark{1},
    Serhat Demirtas\IEEEauthorrefmark{2},
    Ziqiao Wang\IEEEauthorrefmark{2},\\
    Yuhao Jiang\IEEEauthorrefmark{2},
    Jamie Paik \IEEEauthorrefmark{2},
    Kasper Stoy \IEEEauthorrefmark{1},
    Andrés Faíña\IEEEauthorrefmark{1}
    \\
    \\
    }
    \linebreakand
    \IEEEauthorblockA{
        \IEEEauthorrefmark{1}
        REAL Lab, IT University of Copenhagen (ITU)\\
        Email:\mailto{\{baid, rodr, ksty, anfv\}@itu.dk}
    }
    \and
    \IEEEauthorblockA{
        \IEEEauthorrefmark{2}
        RRL, École polytechnique fédérale de Lausanne (EPFL)\\
        Email: \mailto{\{serhat.demirtas, ziqiao.wang, yuhao.jiang, jamie.paik\}@epfl.ch}
    }
}

\maketitle

\begin{abstract}

Object manipulation is a fundamental challenge in robotics, where systems  must balance trade-offs among manipulation capabilities, system complexity, and throughput. \Ac{DMS} use the coordinated motion of actuator arrays to perform complex object manipulation tasks, seeing widespread exploration within the literature and in industry. However, existing DMS designs typically rely on high actuator densities and impose constraints on object-to-actuator scale ratios, limiting their adaptability.
We present a novel \ac{DMS} design utilizing an array of 3-DoF, origami-inspired robotic tiles interconnected by a compliant surface layer. Unlike conventional DMS, our approach enables manipulation not only at the actuator end effectors but also across a flexible surface connecting all actuators; creating a continuous, controllable manipulation surface. We analyse the combined workspace of such a system, derive simple motion primitives, and demonstrate its capabilities to translate simple geometric objects across an array of tiles. By leveraging the inter-tile connective material, our approach significantly reduces actuator density, increasing the area over which an object can be manipulated by $\times1.84$ without an increase in the number of actuators.
This design offers a lower cost and complexity alternative to traditional high-density arrays, and introduces new opportunities for manipulation strategies that leverage the flexibility of the interconnected surface.

\end{abstract}


\section{Introduction}

\acresetall

Object manipulation tasks are commonplace both within the literature and in industry. Systems with few \ac{DoF}, such as conveyors, offer high throughput and payload capacity but are limited in their manipulation capabilities, typically to single-axis translation, lacking the capabilities for rotation or independent object manipulation. In contrast, more dexterous systems like 6-\ac{DoF} robotic arms provide greater precision and control but at the cost of increased complexity, reduced throughput, and limited parallel manipulation capabilities. Such systems also struggle to handle heterogeneous, irregularly shaped, or soft objects without specialized grippers.

\Ac{DMS} offer an alternative approach to manipulation, able to perform precise, multi-dimensional object manipulation.
These systems typically utilize planar arrays of actuators.  These actuators—individually limited to few \ac{DoF}—do not directly grip objects but rely on the coordinated application of force from many actuators to move objects. These systems have been able to perform tasks such as translation, in-plane and out-of-plane rotation—via collective motion. Demonstrated at both the microscopic and macroscopic scales, these actuator arrays offer an appealing alternative to traditional system architectures, offering complex manipulation capabilities, with inbuilt redundancy and high bandwidth afforded by their many cooperating actuators.  %
Actuation mechanisms vary widely, including pistons \cite{xueArrayBotReinforcementLearning2023}, wheels \cite{parajuliActuatorArrayManipulation2014, murpheyFeedbackControlMethods2004}, parallel mechanisms \cite{thompsonRobustPlanarTranslations2021, patilLinearDeltaArrays2023}, and cilia \cite{yimTwoApproachesDistributed2000}. %

In this work, we present an array of 3-\ac{DoF} origami-inspired parallel mechanism robotic tile actuators, interconnected by a layer of flexible material to form a continuous manipulation surface. Unlike conventional \ac{DMS}, this architecture enables actuation both at the end effectors and across the deformable surface. This connective material allows for a far lower actuator density than comparable \ac{DMS}, lowering both the cost and complexity. We analyse the workspace of such a connected multi-actuator system and develop coordinated motion primitives. We demonstrate how the use of continuous compliant surfaces facilitates the manipulation of objects significantly smaller than the inter-actuator spacing, relaxing traditional scale and density constraints as well as unlocking new manipulation strategies.

The main contributions of this work are as follows.
\begin{itemize}
  \item Design of a continuous manipulation surface using 3-\ac{DoF} origami-inspired actuators interconnected through a flexible layer.
  \item Characterization of the system workspace and demonstration of coordinated motion primitives.
  \item Experimental validation of object manipulation of diverse geometries, including objects smaller than the inter-actuator spacing.
\end{itemize}

\begin{figure*}[t]
    \centering
    \includegraphics[width=\textwidth, trim={0 0 0 0},clip]{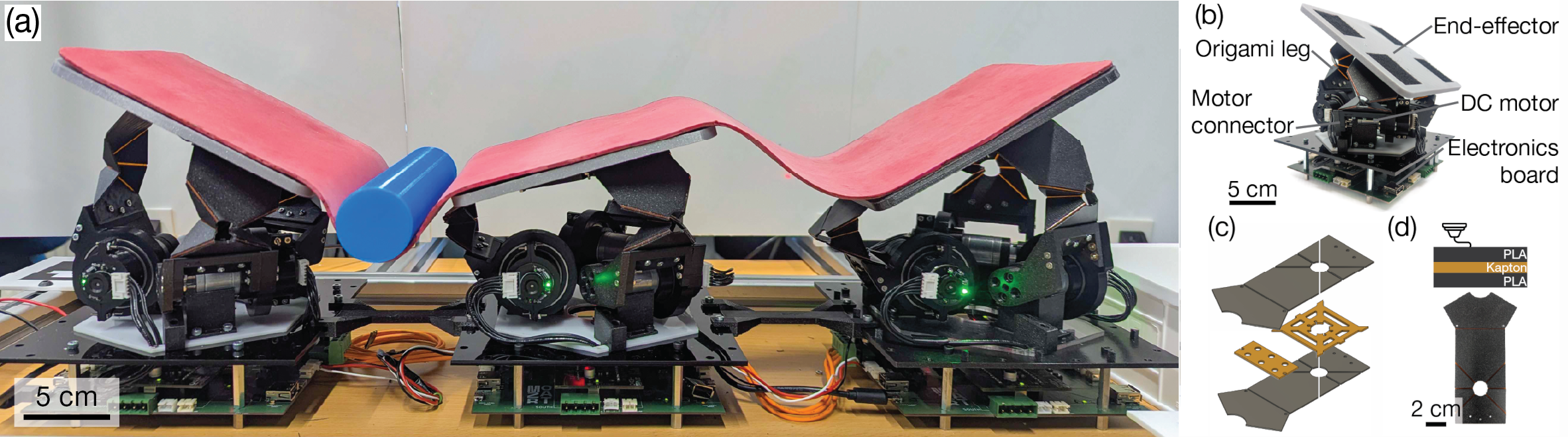}
    \caption{Overview of the robotic surface manipulation system and its components. (a) Actuator array of 3 robotic tiles with an inter-tile distance $D=240\:mm$ and $L=100\:mm$ manipulating a cylinder. (b) A robotic tile module with three origami legs and attachment points for the flexible surface on its end effector. (c) Exploded view of origami leg. (d) Fabricated origami leg composed of $0.127\:mm$ polyimide embedded within $1.2\:mm$ \ac{PLA}.}
    \label{fig:rubber_three_tiles}
\end{figure*}

\section{Related Work}

Various designs of \ac{DMS} have been explored for object manipulation, typically structured as 2D actuator arrays enabling in-plane movement. Some systems, such as those using pistons or wheels, rely on contact with a limited subset of actuators at a time. While this provides planar control, performance degrades when object dimensions fall below the inter-actuator spacing. Conversely, systems using micro-scale actuators (such as MEMS oscillators) utilize a high actuator density, approximating programmable vector fields. Dense arrays of small actuators facilitate the manipulation of objects of varied sizes, though the complexity of control rises for each actuator. 
As such, these systems often sacrifice independent actuator control by grouping actuators and controlling them as a single unit for simplicity \cite{bohringerSensorlessManipulationUsing1995, atakaDesignFabricationOperation2009}. 

A key performance factor is the ratio of object size to actuator pitch (the centre-to-centre spacing between adjacent actuators, or the distance between envelopes of actuators with multiple  \ac{DoF}). Precise control becomes difficult when this ratio is low, with performance deteriorating into inoperability as objects smaller than the spacing fail to make stable contact with multiple actuators. Consequently, most systems manipulate objects much larger than their actuator pitch, restricting the range of manipulable objects or requiring very dense actuator arrays.

While typical \ac{DMS} predominantly move objects substantially larger than individual actuators, actuator-scale manipulation has also been demonstrated; such as macroscopic 3-\ac{DoF} cilia actuators capable of transferring a ball \cite{yimTwoApproachesDistributed2000}. This system required actuators capable of movement in many \ac{DoF} and a highly dense array to facilitate motion. This illustrates the inherent trade-off between system complexity and actuator density when pursuing manipulation at this scale.

Soft connective materials have also been explored to aid manipulation in \ac{DMS}.
Piston-based systems with actuators connected by compliant surfaces create a continuous contact layer, enabling the handling of smaller objects — an inherent advantage of connected actuators \cite{festoWaveHandling2013, hashemControlSoftbodiedXY2016}.
While these systems leverage soft materials to enable novel manipulation strategies, they remain dependent on high actuator density, increasing control complexity and limiting the exploration of manipulation techniques driven purely from interactions with the flexible material. Furthermore, the restricted \ac{DoF} of their linear actuators imposes constraints on overall manipulation capabilities.

Ingle \etal \cite{ingleSoftManipulationSurface2024} explored the limits of a connected piston-based \ac{DMS} in which a soft material suspended between linear actuators serves as the sole end-effector. Their results demonstrate that positional control can be achieved solely through interaction with this soft material, and with a significantly reduced actuator density.

Actuated surfaces with reconfigurable topology have also been investigated for object manipulation \cite{salernoOriPixelMultiDoFsOrigami2020, wangSurfaceBasedManipulation2025, jiangCPGBasedManipulationMultiModule2025}. These systems were shown to manipulate a wide range of objects, including soft and irregular geometries. Their 3-\ac{DoF} actuators share many similarities with our work, featuring square end effectors and comparable workspaces. This research, however, seeks to extend the functionality of these systems by significantly reducing actuator density while broadening their object manipulation capabilities.

\section{System Design}

\subsection{Mechanical design}

This system is comprised of an array of 3-\ac{DoF} parallel mechanism robotic actuators, \cref{fig:rubber_three_tiles}, each incorporating 3 origami motor linkages inspired by the Canfield mechanism \cite{canfieldDevelopmentCarpalWrist1997, zhangExtensibleContinuumRobot2016, salernoNovel4DOFOrigami2016}. We will refer to each of these as a tile. Each tile has a square end effector ($150 \times 150$\si{\milli\meter}) that is linked to the base plate by 3 foldable origami linkages, which are inspired by the waterbomb origami fold. Each origami linkage is driven by a FIT0522 Geared DC motor (DFRobot).

The tiles' origami linkages follow an interpretation of a design showcased by Salerno \etal \cite{salernoOriPixelMultiDoFsOrigami2020}. The linkages are created using $0.165 mm$ Polyimide tape embedded within $1.2mm$ thick \ac{PLA} pieces created via Fused Deposition Modelling. The polyimide is cut into a net-like pattern, shown in \cref{fig:rubber_three_tiles}(c), using a laser cutter. The fabrication of the \ac{PLA} piece is paused partway through to allow for the polyimide net to be properly positioned. 
Polyimide first bonds to the \ac{PLA} through its adhesive coating, but the two materials are ultimately adhered as the \ac{PLA} layers fuse through channels cut into the Polyimide net, embedding it within the structure. The Polyimide is flexible and functions as hinges within the origami piece.

In the modular manipulator array, individual tile modules are interconnected using a flexible but non-stretchable layer, forming a continuous surface. We use a $1.5mm$ thick Linatex\textregistered natural rubber, to which a thin $\SI{0.05}{\milli\meter}$ \ac{PET} film surface layer, used to reduce the friction between the surface and manipulated objects, is adhered using an acrylic adhesive. Due to the masses of objects manipulated, the torques applied by the motors, and the elastic modulus of the carrier rubber ($2 MPa$), the surface stretches very little during normal use and so we consider the material to be inextensible. However, its slight compliance can prevent damage to the modules in the case of hyper-extension. 
The soft material was attached to module end effectors by 4 connective patches of Velcro\textregistered hook-and-loop for ease of fixation and removal. Connection patches were recessed into the end effector to prevent their presence from causing an uneven manipulation surface.

\subsection{Electronics, Communications, and Control}

The system is designed to work as \ac{DMS} consisting of collaborating robotic tiles. Therefore, principles of  modularity and distributed control were integral to the design of electronics and communication systems, to allow for the system configuration to be adapted as necessary with ease.
Each tile is controlled by an onboard Teensy \textregistered MicroMod microcontroller, which controls end-effector positioning and communications between tiles. Each motor is controlled by a DRV8833 motor driver, which attaches to a custom-made main electronics board via a shield.

The tile modules use a serial communication protocol to communicate locally, able to communicate with neighbouring tiles in each of their 4 cardinal directions via connectors linked to the Teensy hardware serial ports. A serial communication protocol allows for tiles to communicate with and  forward commands to their neighbours, either directionally or based upon unique tile IDs. 
Power can be daisy-chained from one module to another using any of the 4 power connector dropouts in each of the tile's cardinal directions, enabling user-friendly cable routing. Each module can allow up to a maximum of $26A$ current through, meaning up to 7 tiles can be connected serially with only a single power connection. 

The system incorporates a cascaded discrete-time velocity-position PID controller for end effector control, ensuring both accurate position tracking as well as the desired end effector manipulation speeds.

\subsection{System configuration}

The system is a \ac{DMS}, constituting an array of robotic modules connected by a soft actuated surface. In this work, we demonstrate a simple linear configuration using 3 tiles equally spaced along a singular axis, as shown in \cref{fig:rubber_three_tiles}. The distance from tile centre to tile centre, $D$, dictates possible values for other system parameters such as inter-tile material length, $L$, which in turn constricts the possible motions of the tiles. We investigate a range of values for $D$ and $L$, and their impact on achievable motions and object manipulation capabilities of the system.

\section{Kinematics and Workspaces}

To analyse the movement of the connected tile array, we first examine the motion of a single tile. Then, we assess how the connective material alters the shared workspace of the array.

\subsection{Kinematics}

\begin{figure}[t]
    \centering
    \includegraphics[width=\columnwidth]{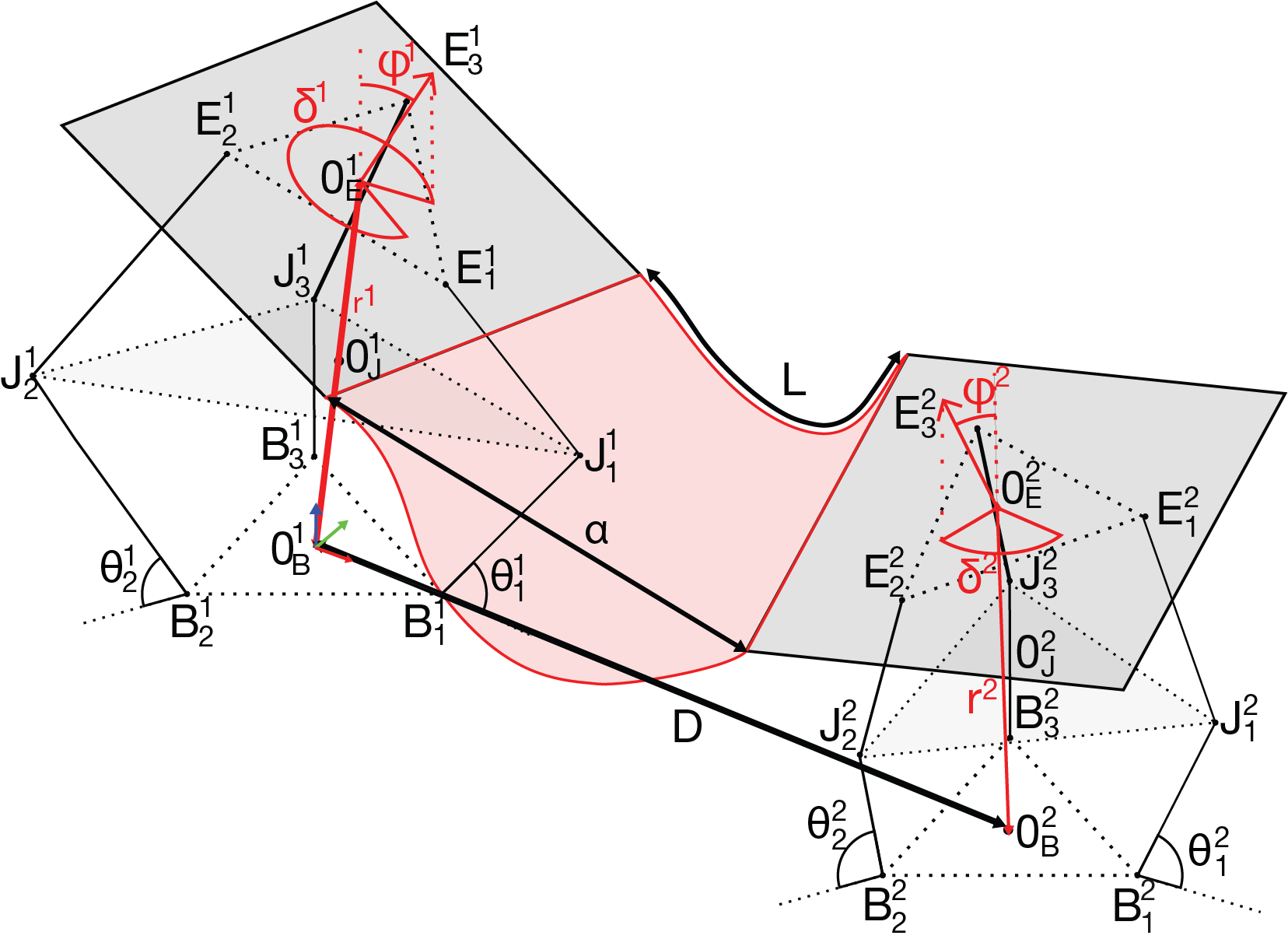}
    \caption{Illustration of two modules modules connected by connective material. This shows motor angles $\theta_i$, which orient the tile into a pose, $(\delta^i, \phi^i, r^{i})$. Modules are at an inter-module distance, $D$, and are connected by a material of length $L$.}
    \label{fig:double_tile_illustration}
\end{figure}

The workspace of the parallel mechanism tiles used in this work depends primarily on two geometric parameters: leg length and leg positioning. The length of the origami legs, $l$, used in this system is divided  equally by the central waterbomb joint (ball and socket), creating the  lower and the upper leg. The base of each of the origami legs is given in polar coordinates $(R, \phi_i, 0)$ with origin at the centre of the robot's base, where $\phi_i$ is the angle of rotation about the Z axis and $R$ is the radius of the circle inscribed by the legs.
In our system, leg length $l = \SI{130}{\milli\meter}$, leg positions $\Phi_1 = \frac{\pi}{3}$, $\Phi_2 = \pi$, $\Phi_3 = \frac{5\pi}{3}$, and the radius of the inscribed circle $R = \SI{44.01}{\milli\meter}$.

The kinematics for a 3-DoF Canfield parallel mechanism have been explored previously, with Mete \etal \cite{meteClosedLoopPositionControl2021} notably deriving the end effector positions as functions of the angles between each leg and the base. Their formulation is directly applicable to our system and is therefore adapted here.

The kinematics of the parallel mechanism take the input leg angles, $\theta_i$, yield a specified end-effector centre $O_E$ relative to the centre of the robot's base $O_B$ where we define the local coordinate frame $XYZ$. The relative position of $O_E$ is expressed in polar coordinates $(r, \phi, \delta)$, where $\delta$ is the angle of rotation about the Z axis (yaw), and $\phi$ is the angle of rotation about the transformed Y axis (composite pitch and roll), which determines end effector tilt. This is illustrated in \cref{fig:double_tile_illustration}. The value of $r$ defined as

\begin{equation}
||\overrightarrow{O_BO_E}|| = r
\end{equation}

\begin{equation}
\overrightarrow{O_BO_E} = r 
\begin{bmatrix}
\sin \phi \cos \delta \\
\sin \phi \sin \delta \\
\cos \phi
\end{bmatrix}
= r \vec{n}
\end{equation}

Where $\vec{n}$ is the unit vector in the direction of $\vec{O_BO_E}$.

Given the system topology, the position of the revolute joints connecting the origami couplings to the base, $B_i$, and to the end effector, $E_i$, are symmetric about a virtual plane passing through the central joint of each origami leg, $J_i$. Therefore, if the position of the central joints is known, the end effector position can be calculated. We use a method described by Mete \etal \cite{meteClosedLoopPositionControl2021} to calculate the locations of the joints, $J_i$, and subsequently the end effector position, $O_B$, for any given leg positions, $\Phi_i$, and angles, $\theta_i$.

\subsection{Workspace - Single Tile}

Using the forward kinematics, we map the reachable workspace for each module. The angles of the legs, $\theta_i$, are constrained to the range $0\le \theta_i\le  \frac{7\pi}{18}$,reduced from the maximum theoretically achievable, $\frac{\pi}{2}$, to prevent damage to the mechanism caused by overextension.

\begin{figure}[t]
    \centering
    \includegraphics[width=\columnwidth]{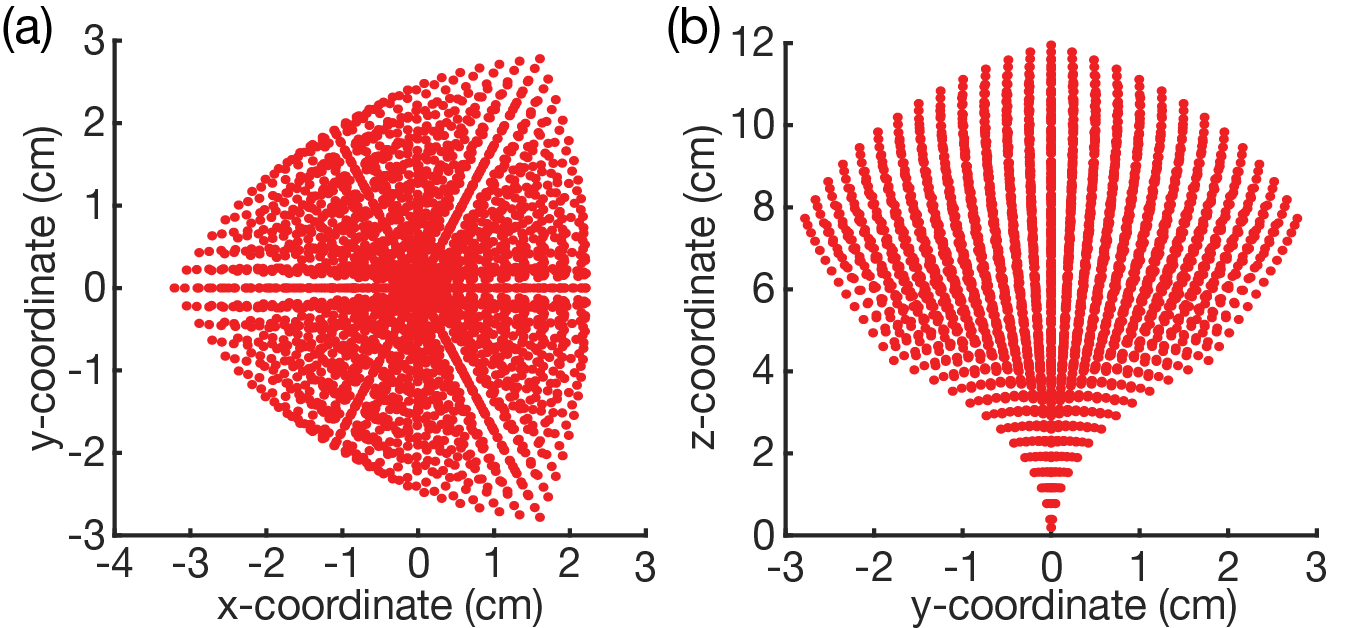}
    \caption{Single robotic tile workspace.}
    \label{fig:workspace}
\end{figure}

\Cref{fig:workspace} shows a visualization of a single tile's workspace, created by sweeping through the range of each of the leg angles and scattering the calculated end-effector position.

\subsection{Workspace - Connected Tiles}

While each tile can independently reach any position in its workspace, this system operates as an array of actuators connected together by a material. As such, we must consider if connected tiles can simultaneously reach points in their individual workspaces while connected by a length of material, $L$.

Given two tiles, $X$ and $Y$, we define two local coordinate frames  at their bases, $O_B^X$ and $O_B^Y$, 
and at their end effectors $O_E^X$ and $O_E^Y$\. The relative position of the frames is determined by each tile's pose. The distance between these two tiles, $D$, is given by $D = ||\overrightarrow{O_B^XO_B^Y}||$.

Each tile has a square end effector with edge length $E_w$, and height $E_h$ whose corners are given in a local frame by:

\begin{equation}
    {C}_F = [( \frac{i \cdot E_w}{2}, \frac{j \cdot E_w}{2}, E_h )]_{O_E^F}, \quad i, j \in \{-1, 1\}, F \in \{X, Y\}
\end{equation}

Considering the position of these corners in a shared frame, the distance between all corners for two robotic modules at any pose can then be calculated.

We consider tiles to be connected at their closest proximal edges. The distances between each pair of corners from each respective tile are calculated, allowing the closest two adjoining edges to be found. With this, the most distal corners of the closest adjoining edge are taken to be the maximum distance between the two robot end effectors. We shall call this distance $\alpha$, illustrated in figure \cref{fig:double_tile_illustration}.

We may then impose a constraint $L\ge\alpha$, where $L$ is the length of the material connecting the two tiles. Configurations that violate the constraint are impossible to reach without straining the connective material. If we consider this impossible (or else undesirable), any pose that violates this constraint is excluded. Thus, the shared reachable workspace is defined as the set of end effector positions that satisfy this constraint. This procedure can be generalized to any number of adjacent connected tiles, creating a shared workspace for the entire array.

It should be noted that distance $\alpha$ measures from end-effector corner to end-effector corner, while D is given from tile centre to tile centre. This means, for a given $E_w$, it is possible for a configuration where $D\ge \alpha$ to be valid.

\begin{figure*}[t]
\centering %
\includegraphics[width=0.9\textwidth, trim={0 0 0 0}, clip]{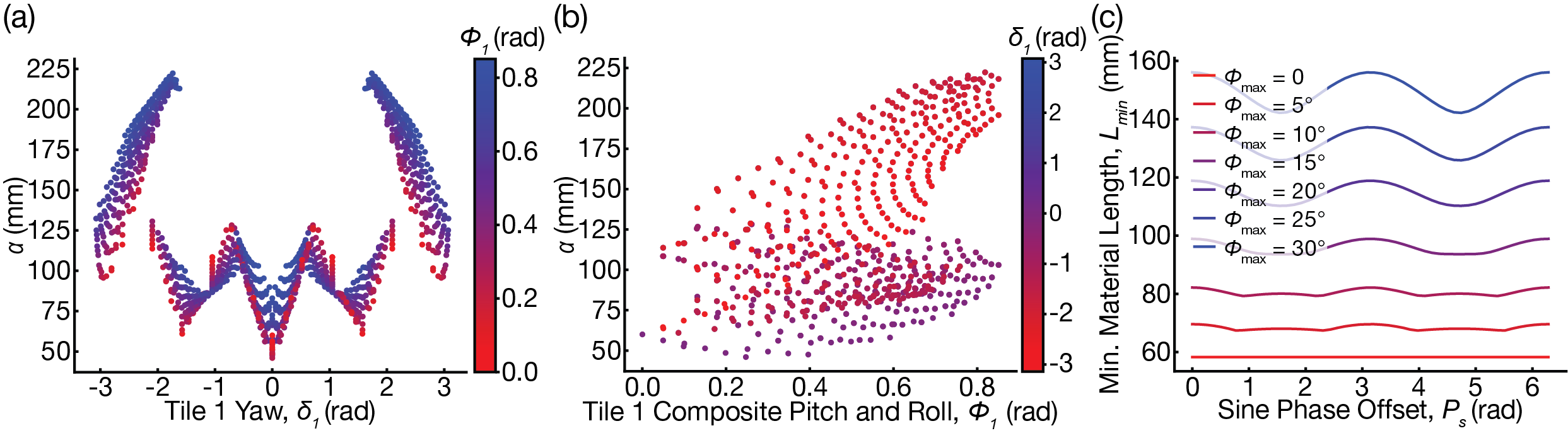}
\caption{Distance between connected end-effectors for static and dynamic tiles. (a, b) Variation of $\alpha$ between two connected tiles as Tile 1 orientation ($\delta_1$, $\phi_1$) changes, with Tile 2 fixed at $\delta_1=0$, $\phi_1=0$, $r_1=\SI{70}{\milli\meter}$ at (210, 0, 0),\si{\milli\meter}, and Tile 1 at (0, 0, 0),\si{\milli\meter} with $r_1=\SI{70}{\milli\meter}$. (b) Effect of phase offset $P_{s}$ on minimum material length $L_{\min}$ under sinusoidal motion at different maximum tilts $\phi_{\max}$.}
\label{fig:delta_phi_alpha_sine_min_length}
\end{figure*}

\Cref{fig:delta_phi_alpha_sine_min_length}(a) illustrates how $\alpha$ varies as $\delta$ and $\phi$ change for two connected tiles at a fixed $r_0$.The plot reveals $\delta$ and $\phi$ have a complex, non-linear effect on $\alpha$.

\section{Movement Pattern Analysis}

Object manipulation depends on tile motion. However, the interconnection of actuators means not all poses can be reached simultaneously,  restricting the shared workspace and limiting feasible motions. In this work, we investigate a simple sinusoidal movement pattern. Here we will investigate how alterations to system parameters such as $D$ and $L$ influence available motions.

\subsection{Sinusoidal Movement Pattern}

For this movement pattern, $\delta$ is fixed for all tiles, aligned with the tile array, an angle we shall call $\Delta$, where $\Delta = 0$ for our system configuration. The value of $\phi$ for each tile oscillates in the range $\phi_{max}\ge\phi \ge -\phi_{max} $, where $\phi_{max}$ is determined by the shared workspace, discussed later. The value for a specific $\phi_i$ at a given time, $t$, is denoted $\phi_{i}^{t}$:

\begin{equation}
    \phi_{i}^{t} = \phi_{max} \sin(2\pi f_{s} t + p_{i}^{s})
\end{equation}

where $f_{s}$ is the frequency of oscillation, and $p_{i}^{s}$ is the desired phase shift for the specific tile.

 A value for base to end effector distance is chosen,  $r_0$. The height of the module during motion is then adjusted by a second sinusoidal pattern. This gives a value for distance from base to end effector, $r$.

\begin{equation}
    r = r_0 + h_{max} \sin(2\pi f_{h} t + p_{h})
\end{equation}

where $h_{max}$ is the maximum height adjustment, $f_{h}$ is the frequency of the height adjustment, and $p_{i}^{h}$ is the phase shift for that tile.

The movement pattern is therefore a tilting tile that oscillates at a set frequency, that may also rise and fall by a given height.

Each tile in the array moves with the same $f_{s}, f_{h}, \phi_{max}, r_0$ and $h_{max}$. However, each tile moves with individual phase offsets, $p_{i}^{s}$ and $p_{i}^{h}$. The phase difference between adjacent tiles for both sinusoids is a constant. This gives values of the phase difference between adjacent tiles, $P_s$ and $P_h$. 

\begin{align}
p_s^i  =  p_s^{i-1} + P_s\\  p_h^i = p_h^{i-1} + P_h  
\end{align}

We impose the constraint that at $t = 0$ the first tile in the array will have $p_s = 0 $. This means that the tile will start level, for any object to be manipulated to rest on top.

Therefore, for tile in the array, $i$,  their pose at a time t is given by:

\begin{equation}
    \begin{bmatrix}
        \delta_i\\ \phi_i\\ r_i
    \end{bmatrix}
    = 
    \begin{bmatrix}
        \Delta \\
        \phi_{max} \sin(2\pi f_{s} t + P_{s}\times(i-1)) \\
        R_0 + h_{\max} \sin(2\pi f_{h} t + P_{h}\times(i-1)) + p_{h}^{i}
    \end{bmatrix}
\end{equation}

\subsection{Achievable Motions}

With the movement pattern defined, we now examine how system parameters \(D\) and \(L\) influence valid motion patterns, including the maximum tilt \(\phi_{\max}\) and the inter-tile phase offset \(P_s\). Since the motion is cyclic, validity is determined if for all time steps in the period it does not violate the material length constraint, $L\ge \alpha$. As tiles are connected in series, performing similar motions with a constant phase offset between neighbours, it is sufficient to evaluate the connection between two adjacent tiles to ensure compliance throughout the array. This approach allows us to determine the minimum length of connective material \(L_{\min}\) required for the constraint to be valid for all time steps across the array when executing a specific motion pattern.

To investigate the effects of changing parameters, we must first start with baseline values. Default values for system and movement parameters were set to the following:

\begin{equation}
\text{Default} =
\begin{Bmatrix}
O_B^1 = (0,0,0)\,\si{\milli\meter}, & O_B^2 = (400,0,0)\,\si{\milli\meter}, \\
f_s = \SI{1}{\hertz}, & f_h = \SI{1}{\hertz}, \\
h_{\max} = \SI{0.0}{\milli\meter}, & \phi_{\max} = \tfrac{\pi}{9}, \\
P_s = \SI{0}{\hertz}, & P_h = \SI{0}{\hertz}, \\
r_0 = \SI{70}{\milli\meter}
\end{Bmatrix}
\end{equation}

Unless otherwise stated, the parameters in the following examples are set to the default.

As an initial test, we investigate how $L_{min}$ varies with changes in $D$. This is done by positioning the second tile at \((x, 0, 0)\) and sampling 50 evenly spaced values of \(x\) within the range \([210,  400]\,\si{\milli\meter}\). \(L_{\min}\) increases linearly with inter-tile distance, as expected. Next, we investigate the effect of $P_{s}$ on $L_{min}$.  We sampled 50 values for $P_s$ in the range $[0, 2\pi]$.  The value for $L_{min}$ varies sinusoidally with phase offset. Finally, we investigate the effect the value for $\phi_{\max}$ has on $L_{\min}$. We sampled 50 values for $\phi_{\max}$ in the range $[0, \frac{\pi}{6}]$, from which the value of $L_s$ was found to vary non-linearly. \Cref{fig:delta_phi_alpha_sine_min_length}(b) shows a visualization of how $L_{min}$ varies with changes to $\phi_{\max}$ and $P_s$.

\section{Object Manipulation}

This section examines the system's capability to perform object manipulation, specifically linear object translation across the array. 

\subsection{Experimental Setup}

Objects were placed on top of the centre of the left-most tile in the array. The task was to translate objects using tile motion until they reached the edge of right-most tile, falling into a detection area. If an object was able to reach the detection area within a threshold time, set heuristically at $20s$, the manipulation was successful. All experiments were repeated at least 3 times, with 3 consecutive successes required for a manipulation strategy to be considered successful.

Object positions were tracked with a 2-camera HSV based tracking system using the OpenCV library\cite{bradskiOpencvLibrary2000}. To facilitate robust visual detection, objects were created using a bright, mono-coloured \ac{PLA}. Aruco markers fixed at known locations were used to translate position in pixel space to real-world spatial coordinates.

\subsection{Translation Experiments - Rolling}

To evaluate the system's capability of manipulating objects that roll, we tested manipulation of a cylinder($\diameter40 \times 170\si{\milli\meter}$).
Building upon the previously described sinusoidal motion pattern, through examination of the effects of varying parameters, we created two distinct movement patterns. We then experimentally validated object translation capabilities under fixed values of $L$ and $D$.

Translation requires an object to move from a tile onto the inter-tile material and subsequently onto the neighbouring tile. With the sinusoidal motion pattern, translation was ineffective when tiles were either in phase or completely out of phase ($P_s = 0, \pi$), as no net directionality was produced. The frequency of the oscillation had to ensure sufficient time for the object to roll fully into the inter-tile region before the next cycle begins. Experimentally, $f_s=\SI{0.25}{\hertz}$ with $P_s = \tfrac{\pi}{2}$ provided the most effective translation in this regard.

Tilting readily moves an object onto the inter-tile material, but transferring to the next tile while it is in a receptive pose is more challenging. As tiles rotate about their base, not the centre of their end effector, the inter-tile material becomes slack as tiles tilt towards each other, and becomes taut when tilting away. During motion, objects often move between these areas of slack material before being pushed out if the material becomes taut. By using our kinematics analysis, we determine the values of $\phi$ and $r$ that will ensure tautness in the material for a given tile spacing and inter-tile material length.

Height oscillations with a phase offset between neighbouring tiles create a gradient for the cylinder to roll down. These oscillations also increase end-effector separation, $\alpha$, thereby pulling the inter-tile material taut.

\begin{equation}
\mathbf{A} =
\begin{Bmatrix}
f_s=0.25Hz, f_h = 0.25Hz , P_s = \frac{\pi}{2}, P_h = \frac{\pi}{2}
\end{Bmatrix}
\label{eq:movement_pattern_a}
\end{equation}

\begin{equation}
\mathbf{B} =
\begin{Bmatrix}
f_s=0.25Hz, f_h = 0.5Hz, P_s = \frac{\pi}{2}, P_h = 0
\end{Bmatrix}
\label{eq:movement_pattern_b}
\end{equation}

 \begin{figure}[t]
\centering %
\centering
\includegraphics[width=0.9\columnwidth, trim={0 0 0 0}, clip]{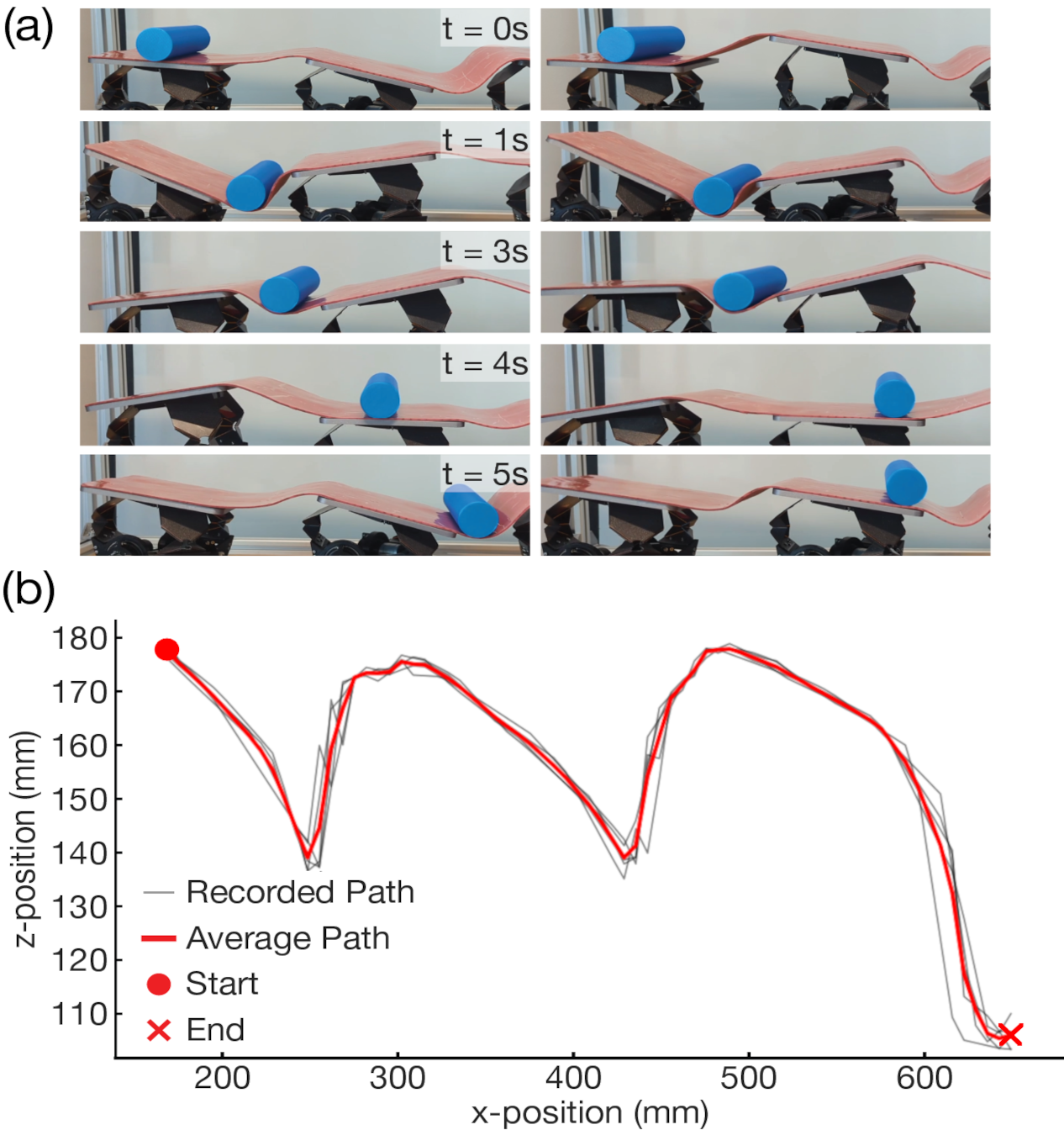}
\captionsetup[figure]{skip=4pt}
\caption{Cylinder manipulation experiments. (a) Video frames showing a cylinder manipulated by tiles, illustrating the similarity in object motion between the two patterns. (b) Average trajectory of the cylinder over five trials when using pattern B with $D = 240\si{\milli\meter}$ and $L = 100\si{\milli\meter}$.}
\label{fig:fig6combined}
\end{figure}

From these requirements, we identified two movement patterns, denoted $\mathbf{A}$ and $\mathbf{B}$, given in \cref{eq:movement_pattern_a,eq:movement_pattern_b}. As illustrated in \cref{fig:fig6combined}, both patterns roll the cylinder into the inter-tile material and then eject it in the direction of translation as the material is pulled taut. This ejection results from the first tile tilting away and a height offset due to $P_h$, both increasing $\alpha$. The receiving tile begins tilted toward the object and levels as the object is transferred onto it.

Allowing tiles to tilt without restriction, thereby enabling a larger achievable $\phi_{\max}$, was experimentally found to improve translation. This occurs either when the phase difference $P_s$ was near zero - though this conflicted with the directionality requirement - or when the inter-tile material length $L$ was high relative to the inter-tile distance $D$. However, if $L$ exceeded a threshold, tiles could not pull the material taut to eject objects. In all successful cases, the movement parameters pulling the inter-tile material taut was essential, making $L:D$ a critical design parameter.

 A trade-off exists between achievable $\phi_{\max}$ and $h_{\max}$. Experimental results indicated that maximizing $h_{\max}$ was most advantageous; it was therefore set to its maximum value of $\SI{15}{\milli\meter}$. Subsequently, $\phi_{\max}$ was set to its upper limit as allowed by the workspace, up to a maximum of $\frac{5\pi}{36}$ - to ensure that each tile could achieve the required pose.

The calculated value of $\alpha$ was found to underestimate the achievable distances in the physical system.  This discrepancy arose from a combination of factors such motor backlash, encoder resolution, and compliance in the origami components. To compensate, an additional factor of $\frac{\pi}{18}$ was added to the calculated $\psi_{\max}$ to account for the slack, but not exceeding the maximum of $\frac{25\pi}{180}$. .

\begin{table}[t]
    \captionsetup{skip=2pt} %
    \setcellgapes{2pt}
    \makegapedcells
    \centering
    \caption{Results of cylinder translation experiments. Symbols: {\checkmark} successful translation for both , {\ding{55}} failed translation, Inv invalid configuration (distance $>$ material length), and -- untested.}
    \label{table:manipulation_results}
    \begin{tabularx}{\columnwidth}{ l | X X X X X X X X X }
        L (mm) & & & & & & & & & \\
        200 & - & - & - & - & - & \FailureCell & \FailureCell & \FailureCell & \SuccessCell \\
        150 & - & - & \FailureCell & \FailureCell & \FailureCell & \SuccessCell & \InvalidCell & \InvalidCell & \InvalidCell \\
        100 & - & \FailureCell & \SuccessCell & \SuccessCell & \InvalidCell & \InvalidCell & \InvalidCell & \InvalidCell & \InvalidCell \\
        50 & \FailureCell & \SuccessCell & \InvalidCell & \InvalidCell & \InvalidCell & \InvalidCell & \InvalidCell & \InvalidCell & \InvalidCell \\
        0 & \FailureCell & - & - & - & - & - & - & - & - \\
        \midrule
        D (mm) & 180 & 200 & 220 & 240 & 260 & 280 & 300 & 320 & 340 \\
    \end{tabularx}
\end{table}

 \Cref{table:manipulation_results} summarises the manipulation experiments. Without a connective material ($L=0$) we show the object was unable to translate. By using a interconnecting material and applying the two movement patterns, we successfully translated the object across multiple different inter-tile distances and material lengths, including an inter-tile distance of $\SI{340}{\milli\meter}$ - over twice the width of the tile end effector. With an inter tile distance, D, and an end effector width, w, the effective traversable distance can be approximated by:
 
 \begin{equation}
\text{Increase factor} =
\frac{2D+ w}{3w} \approx 1.84
\end{equation}
 
 This corresponds to an approximate $1.84\times$ increase in traversable distance, along with a proportional decrease in actuator density relative to an equivalent densely packed system, though the true value would depend on factors such as envelope overlap. These results demonstrate the effectiveness of the inter-connective material layer, providing a flexible, continuous surface that circumvents traditional object-to-actuator pitch restrictions, able to manipulate an object much smaller than the actuator end effectors and the inter-actuator spacing.

\subsection{Translation Experiments - Sliding}

To demonstrate the manipulation of sliding objects, we manipulated a flat circular disk ($\diameter \SI{60}{\milli\meter} \times \SI{5}{\milli\meter}$) along the array. The disk translates exclusively by sliding - as such, it required a modified manipulation strategy.

\begin{table}[t]
\centering
\caption{State definitions for sliding object translation sequence. $\Phi_F$  and $\Phi_B$ denote a specific forward and backward tilt angle, while $R_H$ and $R_L$ are set high and low tile heights.}
\begin{tabular}{c | c c c}
\toprule
\textbf{State} & \boldmath{$\delta$} & \boldmath{$\phi$} & \boldmath{$r_0$} \\
\midrule
1 & $0$ & $\Phi_F$ & $R_H$ \\
2 & $\pi$ & $\Phi_B$ & $R_H$ \\
3 & -- & $0$ & $R_H$ \\
4 & $\pi$ & $\Phi_B$ & $R_L$ \\
5 & -- & $0$ & $R_L$ \\
6 & $0$ & $\Phi_F$ & $R_H$ \\
\bottomrule
\end{tabular}
\label{tab:states}
\end{table}

Insights from cylinder manipulation experiments guided the development of a sequence of movements to successfully translate the disk. Transport relies on the same mechanism - capture within the inter-tile material and subsequent ejection through coordinated tile movement. The sinusoidal motion for rolling objects was replaced with a cycling set of 6 states, shown in \cref{tab:states}, that set tile poses, tilting them in either in the direction of travel, against it, or had them level with the ground; all at a specified height.

\begin{figure}[t]
\includegraphics[width=\columnwidth, trim={0 0 0 0}, clip]{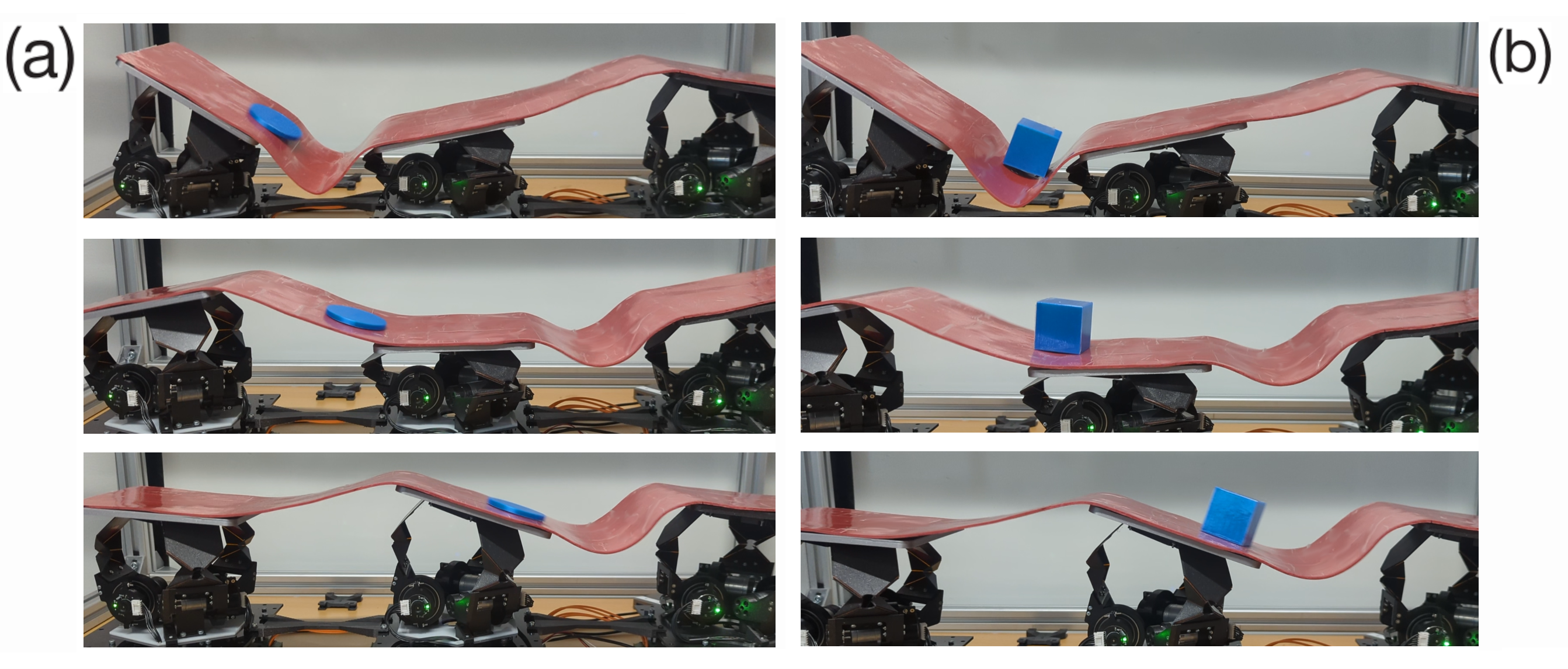}
\caption{Video frames of the first 3 cycles of manipulation experiments. (a) Manipulating the disk (b) Manipulating the cube}
\label{fig:disk_cube}
\end{figure}

The tiles cycle through states sequentially, changing state after a set time period ($t_s$). A value of $t_s = 5s$ was chosen as this ensured objects had sufficient time to traverse to the next part of the array. However, for shorter values of D this could have been reduced. Each tile is always 3 states ahead of their neighbour; as we have 6 states, this means that tiles follow an alternating pattern.

The large contact surfaces of each object caused significantly more friction than was experienced with the cylinder. Therefore, a sinusoidal vibration (frequency = \SI{5}{\hertz}, amplitude = \SI{5}{\milli\meter}) was applied to the tiles $r_0$ to ensure that objects did not become stuck.

\begin{table}[t]
    \setcellgapes{2pt}
    \makegapedcells
    \centering
    \caption{Successful manipulation parameters for state translation of non-cylindrical objects for different material lengths (L) and inter-tile spacings (D)}
    \label{table:state_machine_manipulation_parameters}
    \begin{tabularx}{\columnwidth}{ l | X X X X}
    \toprule
        L:D(mm) & $\Phi_F$ & $\Phi_B$ & $R_H$ (mm) & $R_L$ (mm)\\
        \midrule
        100:240 & $\frac{5\pi}{36}$ & $\frac{\pi}{12}$ & 70 & 40 \\
        150:280 & $\frac{\pi}{6}$ & $\frac{\pi}{12}$ & 80 & 40 \\
        200:320 & $\frac{5\pi}{36}$ & $\frac{\pi}{12}$ & 85 & 40 \\
        \bottomrule
    \end{tabularx}
\end{table}

By cycling these states, both objects across the array with an inter-tile distance of 240, 280, and 340\si{\milli\meter}, with inter-tile material lengths of 100, 150, and 200\si{mm} respectively. The values of the parameters, $\Phi_F, \Phi_B, R_H,$ and $R_L$ were selected to ensure the inter-tile material was pulled taut for different spacings and material lengths, using the calculation of $\alpha$ as described in our kinematics analysis. The specific values are detailed in \cref{table:state_machine_manipulation_parameters}. Greater tile spacing required tiles to tilt further or have a greater difference in height between their high and low state to ensure that the inter-tile material was pulled taut, and that a significant gradient was maintained when traversing the longer connective material between tiles.

\subsection{Translation Experiments - Sliding and Rolling}

Finally, we repeated the translation experiments with a cube ($\SI{40}{\milli\meter}^3$). The cube alternates
between rolling and sliding depending on its orientation relative
to the surface during the actuation. 

Using the same cycling movement pattern used for the disk, we were successful at translating the  object across the same inter tile distances, using the same length of material, and movement parameters.  However, successful translation of the cube was highly dependent on initial positioning, as rolling can initiate unpredictably, producing semi-chaotic motion and occasional failures using this open-loop manipulation.

\section{Discussion}

In this work, we demonstrated object translation using a linear actuator array. A natural extension is to employ a two-dimensional \ac{DMS} array, enabling more complex object manipulations beyond simple linear motion. This extension is feasible and is an avenue for future work. However, computing the valid workspace using the motion of only two connected tiles would not hold when tiles are interconnected in multiple dimensions. The same is true if the tiles were to move not in a simple periodic motion. This can significantly increase the computational burden of determining valid workspaces.

We demonstrated translation with simple geometric objects. While this serves as a proof-of-concept, objects of other geometries may follow different trajectories and require modified movement patterns for reliable translation. Irregular geometries may also move in a less predictable manner, necessitating novel manipulation strategies. Although our focus is on objects smaller than the inter-actuator spacing, larger objects that contact multiple tiles simultaneously would require different manipulation strategies, such as gating. We would also like to investigate the manipulation of deformable objects, for which the absence of grasping may be advantageous.

Our experiments used an open-loop, hand-crafted manipulation strategy to demonstrate the system's functionality and the role of the connective material. Future work should explore closed-loop control and manipulation strategies for greater manipulation precision, enabling targeted motions based on object location and allowing selective application of movements such as vibration when an object becomes stuck.

Here we consider the material as inextensible. This is not true of real-world materials; though this approximation holds for the object masses and motor torques tested. In the real world, compliance may cause hysteresis and energy loss, potentially reducing manipulation precision and repeatability. Future work should examine how material stretch affects object manipulation, reachable workspaces, positional drift over cycles, and motor loading, as well as exploring ways to exploit or mitigate this behaviour.

\section{Conclusion}

This work introduces a soft, interconnected actuator array that enables effective object manipulation with significantly reduced actuator density. By interconnecting actuators with a flexible material the proposed design not only extends the reachable workspace but also relaxes traditional scale and density constraints inherent to distributed manipulator systems. Through kinematic analysis and the implementation of a sinusoidal and a cycling state based movement patterns, the system successfully translated objects of different geometries across a large distance, achieving a $1.84\times$  increase in .traversable distance when compared to a similar, densely packed system that did not use connective material.

\section{Acknowledgments}
This work was conducted as part of the MOZART project, funded by the European Union, EU project id: 101069536.

\bibliography{references}
\bibliographystyle{IEEEtran}
\end{document}